\documentclass{article}

\usepackage{microtype}
\usepackage{graphicx}
\usepackage{subcaption}
\usepackage{booktabs} 

\usepackage{hyperref}

\usepackage{amssymb}
\usepackage{amsmath}
\usepackage{amsthm}
\usepackage{amsfonts}
\usepackage{commath}

\usepackage{graphicx}
\usepackage{subcaption}
\usepackage[export]{adjustbox}
\usepackage{wrapfig}

\usepackage{algorithm}
\usepackage{algcompatible}
\usepackage[noend]{algpseudocode}

\usepackage[colorinlistoftodos]{todonotes}

\usepackage{setspace}

\usepackage[inline]{enumitem}

\newcommand{\LINECOMMENT}[1]{\STATE {\color{brown}\ttfamily\small \(\triangleright\) #1}}

\theoremstyle{plain}
\newtheorem{theorem}{Theorem}[section]

\newtheorem{lemma}[theorem]{Lemma}

\theoremstyle{definition}
\newtheorem{definition}[theorem]{Definition}
\newtheorem{assumption}[theorem]{Assumption}
\theoremstyle{remark}

\newcommand{\alg}{\textsc{Lumos}}

\newcommand{\alggen}{\textsc{Lumos}}

\usepackage[main, final]{neurips_2026}

\usepackage[utf8]{inputenc} 
\usepackage[T1]{fontenc}    
\usepackage{hyperref}       
\usepackage{url}            
\usepackage{booktabs}       
\usepackage{amsfonts}       
\usepackage{nicefrac}       
\usepackage{microtype}      
\usepackage{xcolor}         

\title{Learning to Undo: Transfer Reinforcement Learning under State Space Transformations}

\author{%
  Mridul Mahajan\textsuperscript{1} \quad
  Aldo Pacchiano\textsuperscript{1,2$\dagger$} \quad
  Xuezhou Zhang\textsuperscript{1$\dagger$} \\
  \textsuperscript{1}Boston University \\
  \textsuperscript{2}Broad Institute of MIT and Harvard \\
  \texttt{\{mridulm,pacchian,xuezhouz\}@bu.edu}
}

\begin{document}

\maketitle
\begingroup
  \renewcommand{\thefootnote}{\fnsymbol{footnote}}
  \footnotetext[2]{Equal senior authorship.}
\endgroup

\begin{abstract}
  Transfer learning in reinforcement learning (RL) has shown strong empirical success. In this work, we take a more principled perspective by studying when and how transferring knowledge between MDPs (a source and a target) can be provably beneficial. Specifically, we consider the case where there exists an undo map such that applying this map to the target’s state space recovers the source exactly. We propose an algorithm that learns this map by matching state feature statistics gathered from both MDPs, and then uses it to transfer the source policy to the target. We theoretically justify the algorithm by analyzing the setting when the undo map is linear and the source is linearly-$Q^\star$ realizable, where our approach has strictly better sample complexity than tabula rasa RL in the target MDP. Empirically, we demonstrate that these benefits extend beyond this regime: on challenging continuous control tasks and Atari games, our method achieves significantly better sample efficiency. Overall, our results highlight how shared structure between tasks can be leveraged for efficient transfer of policies across environments. Our codebase is publicly available at \href{https://github.com/MridulMahajan44/neurips2026-learning-to-undo}{\nolinkurl{github.com/MridulMahajan44/neurips2026-learning-to-undo}}.
\end{abstract}
\section{Introduction}
\begin{wrapfigure}{r}{0.413\textwidth}
    \centering
    \vskip -0.4cm
    \includegraphics[
        width=0.42\textwidth,
    ]{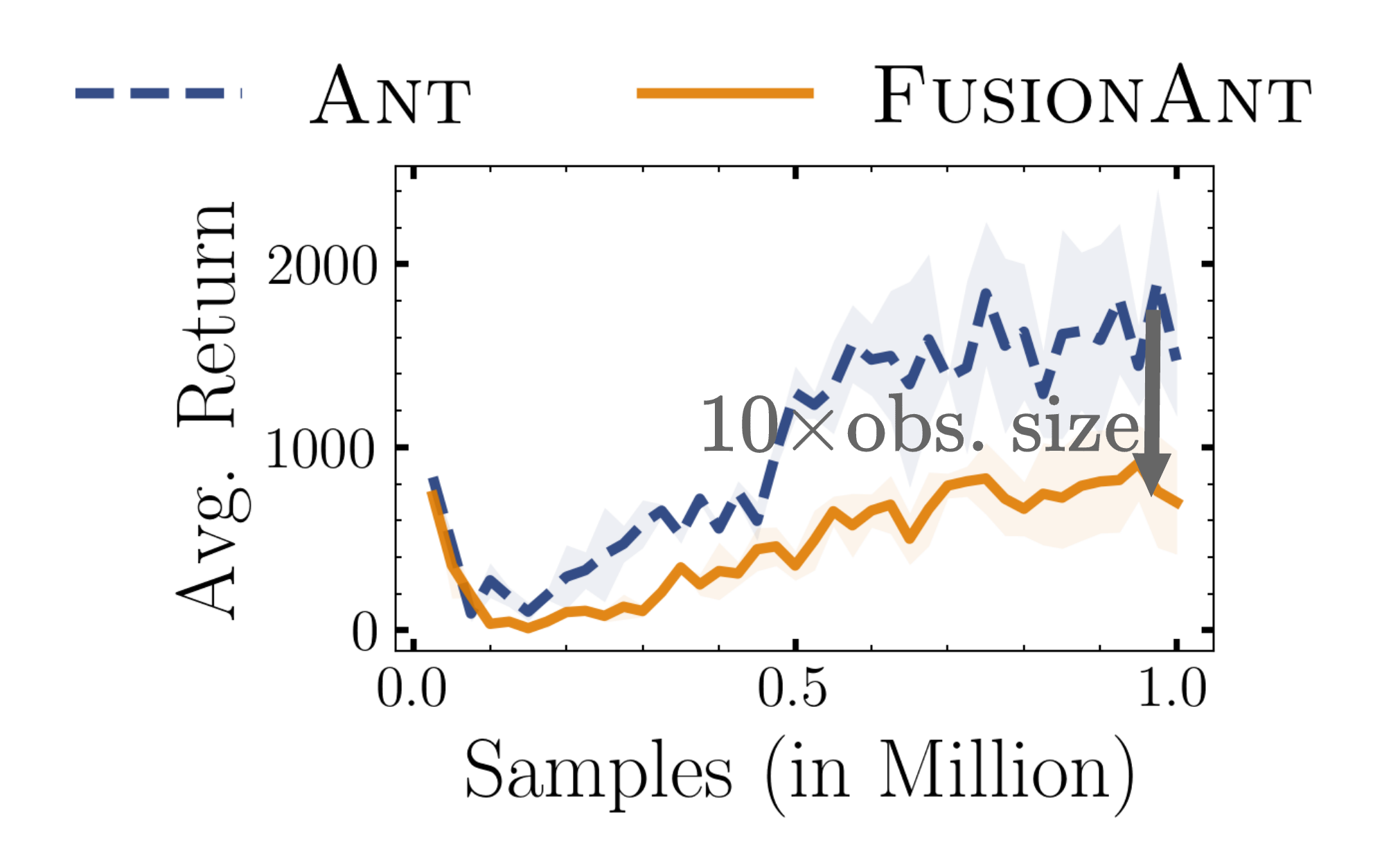}
    \caption{State space transformations can significantly decrease sample efficiency (e.g., increasing observation size by $10\times$).}
    \label{fig:intro_plot}
      \vskip -0.6cm
\end{wrapfigure}

Reinforcement learning (RL) has seen rapid progress in recent years and has achieved strong performance in complex tasks, such as video games, locomotion, manipulation, and navigation \citep{atari, anet, zhu2019dexterous}. Despite this, learning high-quality policies from scratch typically requires millions of environment interactions, which limits the applicability of RL in real-world domains where data collection is costly or time-consuming \citep{challenges}.

Transfer learning aims to overcome this challenge by leveraging knowledge learned from source tasks to accelerate learning on related target tasks \citep{oldtransfersurvey}. It has the potential to drastically improve sample efficiency by effectively reusing prior experience. While transfer learning has seen strong empirical success in RL \citep{transfersurvey}, our goal is not to design a state-of-the-art transfer method that works across arbitrary variations in tasks. Instead, we study the problem from a more principled lens, where we explicitly model the similarity between tasks and exploit it to obtain an algorithm that not only has provable sample complexity improvements over learning from scratch, but also works empirically.

\begin{figure*}[t]
    \centering
    \includegraphics[
        width=\textwidth,
        trim=0cm 2cm 0cm 0cm,
        clip
    ]{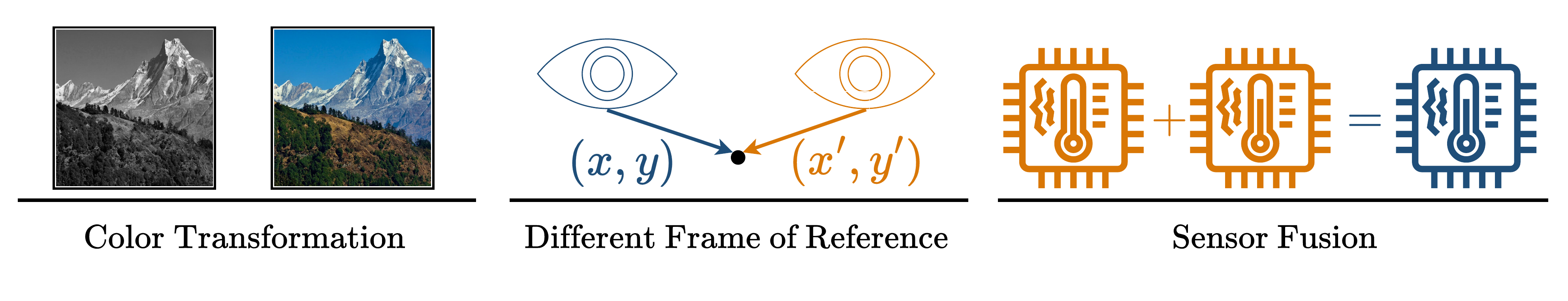}
    \caption{State space transformations are grounded in real-world applications.}
    \label{fig:transformations}
    \vskip -0.45cm
\end{figure*}

An interesting and practically relevant setting with such structure is that of state space transformations: the source and target tasks differ due to transformations in the state space. For instance, real-world transformations such as color transformation from grayscale to RGB, change in frame of reference, and sensor fusion -- all are state space transformations (see Fig.~\ref{fig:transformations}). Such transformations provide a rich and practical setting to design principled transfer learning methods that explicitly model them. One might assume that such transformations are trivial to handle since the underlying task remains the same. However, this intuition is misleading: as shown in Fig.~\ref{fig:intro_plot}, even a $10\times$ increase in observation size in the sensor fusion setting, where readings must be aggregated meaningfully, is enough to render tabula rasa RL severely suboptimal, which highlights that principled transfer is necessary.

State-space and observation transformations naturally occur in many practical RL scenarios where the underlying task dynamics remain unchanged, but the agent’s interface to the environment changes. For instance, in robotics, policies are often trained in simulation and then deployed on real systems, since real-world training is significantly more costly and difficult to parallelize \citep{sim2realsurvey}. In the transition from simulation to the real system, differences between simulated and physical sensors cause discrepancies in sensing and perception, and thus, introduce shifts in the agent’s observation space \citep{sensorfusionrl}. In such settings, relearning policies from scratch can be very sample-inefficient even though the underlying control problem essentially remains the same.

Furthermore, low-dimensional privileged state features are often available during development, but deployment must operate under high-dimensional inputs such as camera observations. Even when both types of observations are available, learning with compact representations is often substantially easier \citep{priviligedrl}, which highlights the utility of learning under state-space transformations.

In this work, we address this setting by learning an \emph{undo map} that transforms the state space in the target back to the source. Instead of learning a policy from scratch, we recover a function that \emph{undoes} the transformation applied to the target states, which allows reusing source policies in the target task.

To this end, we introduce a distributional notion of \emph{undo discrepancy} that measures the quality of a candidate undo map. Minimizing this discrepancy is equivalent to matching the state feature distributions in the source MDP and the corresponding \emph{undone} target feature distributions at each timestep under open-loop action sequences. We then propose a practical transfer learning algorithm that optimizes this objective using samples from the target. Theoretically, we show that when the undo map is linear and the source is linearly-$Q^\star$ realizable, our approach has strictly better sample complexity than learning from scratch.

To empirically evaluate our method, we use challenging continuous control tasks and Atari games, and construct target tasks by applying transformations such as changes in frame of reference and coordinate system, sensor fusion-inspired transformations, and color shifts. We also consider perception gaps such as moving from sensor observations in the source to pixel observations in the target, and moving from semantic bitmaps in the source to pixel observations in the target. Our results show that learning the undo map consistently outperforms learning the target policy from scratch in terms of sample efficiency, where it often achieves near-optimal performance with an order of magnitude fewer samples. These results highlight the practical utility of our approach for transfer RL.

To summarize, our work makes the following contributions:
\begin{enumerate}[label={\Roman*.},leftmargin=*]
    \item We propose a novel transfer learning method, {\alg} (\textbf{L}earning to \textbf{U}ndo by \textbf{M}atching \textbf{O}bservation \textbf{S}tatistics), that performs regression on state feature statistics to \emph{undo} state-space transformations in the target MDP (Sections~\ref{sec:dist_undo}~and~\ref{sec:learn_undo}).
    \item Theoretically, we prove that our method achieves strictly better sample complexity than learning from scratch when the undo map is linear and the source is linearly-$Q^\star$ realizable (Section~\ref{sec:theory}).
    \item We empirically demonstrate the practicality of our method via experiments on challenging continuous control tasks and Atari games involving state-space transformations (Section~\ref{sec:experiments}).
\end{enumerate}
\section{Problem Setup}

\textbf{Tasks.} 
We model each task as a finite-horizon episodic Markov Decision Process (MDP), defined by the 7-tuple $(\mathcal{S}, \mathcal{A}, \mathbb{P}, r, H, \mu, \phi)$, where $\mathcal{S}$ is the state space, $\mathcal{A}$ is the action space, $\mathbb{P}: \mathcal{S} \times \mathcal{A} \to \mathcal{P}(\mathcal{S})$ defines the transition dynamics, $r: \mathcal{S} \times \mathcal{A} \to \mathbb{R}$ is the reward function, $H$ is the fixed horizon length, $\mu \in \mathcal{P}(\mathcal{S})$ is the initial state distribution, and $\phi: \mathcal{S} \to \mathbb{R}^{d}$ is a fixed feature map. W.l.o.g.,  we assume that $d \leq H$. The agent interacts with the task using a non-stationary policy $\pi_h: \mathbb{R}^d \to \mathcal{P}(\mathcal{A})$, and for $0 \leq h < H$, chooses an action $a_h \sim \pi_h(\cdot \mid \phi(s_h))$. The value of the policy $\pi$ is defined as the expected return $V^\pi \triangleq \mathbb{E}_{s_0 \sim \mu, \pi, \mathbb{P}} \left[\sum_{h=0}^{H-1} r(s_h, a_h) \right]$, and the agent's goal is to learn a policy $\pi^\star$ that maximizes this quantity, i.e., $V^* = \max_{\pi} V^\pi$. The state-action value function $Q^\pi(s, a)$ denotes the expected return when taking action $a$ in state $s$, and following policy $\pi$ thereafter.

\textbf{Transfer Learning.} The transfer learning setting follows the protocol in which the agent interacts with a source task $\mathcal{M}_S = (\mathcal{S}^S, \mathcal{A}^S, \mathbb{P}^S, r^S, H, \mu^S, \phi_S)$ to learn a policy $\pi_S$, and then leverages the learned policy to speed up learning in the target task $\mathcal{M}_T = (\mathcal{S}^T, \mathcal{A}^T, \mathbb{P}^T, r^T, H, \mu^T, \phi_T)$. We assume that the agent has complete access to the source MDP $\mathcal{M}_S$, either by knowing its dynamics and reward function exactly or by being allowed to interact with it arbitrarily without incurring any sample complexity cost. In contrast, interactions with the target MDP $\mathcal{M}_T$ are costly. Further, the target MDP does not provide reward feedback during adaptation.

\textbf{Shared Structure and Undo Map.} We assume that both tasks are exactly the same except that the agent observes states through task-specific feature maps: $\phi_S: \mathcal{S} \to \mathbb{R}^{d_S}$ in the source and $\phi_T: \mathcal{S} \to \mathbb{R}^{d_T}$ in the target. These feature maps are related by an unknown transformation (the \emph{undo} map) $U_\star: \mathbb{R}^{d_T} \to \mathbb{R}^{d_S}$, such that for all $s \in \mathcal{S}$, $\phi_S(s) = U_\star \left( \phi_T(s) \right)$. This relationship induces a correspondence between policies: for any policy $\pi_S: \mathbb{R}^{d_S} \to \mathcal{P}(\mathcal{A})$ defined in the source feature space, we define the corresponding policy in the target task as $\pi_T\!\left( \phi_T(s) \right) = \pi_S\!\left( U_\star \left( \phi_T(s) \right) \right)$. These policies are equivalent in terms of their value functions: $V_S^{\pi_S}(s) = V_T^{\pi_T}(s)$ for all $s \in \mathcal{S}$. This allows source policies to transfer to the target despite differing state representations. 

\textbf{Objective.} Given complete access to the source MDP $\mathcal{M}_S$, our objective is to learn a near-optimal policy for the target MDP $\mathcal{M}_T$ using as few samples from the target as possible.
\section{{\alggen}: Learning to Undo by Matching Observation Statistics}
\label{sec:method_general}

\begin{wrapfigure}{r}{0.58\textwidth}
    \centering
    \includegraphics[
        width=0.47\textwidth,
        trim=0.5cm 0cm 17cm 0cm,
        clip
    ]{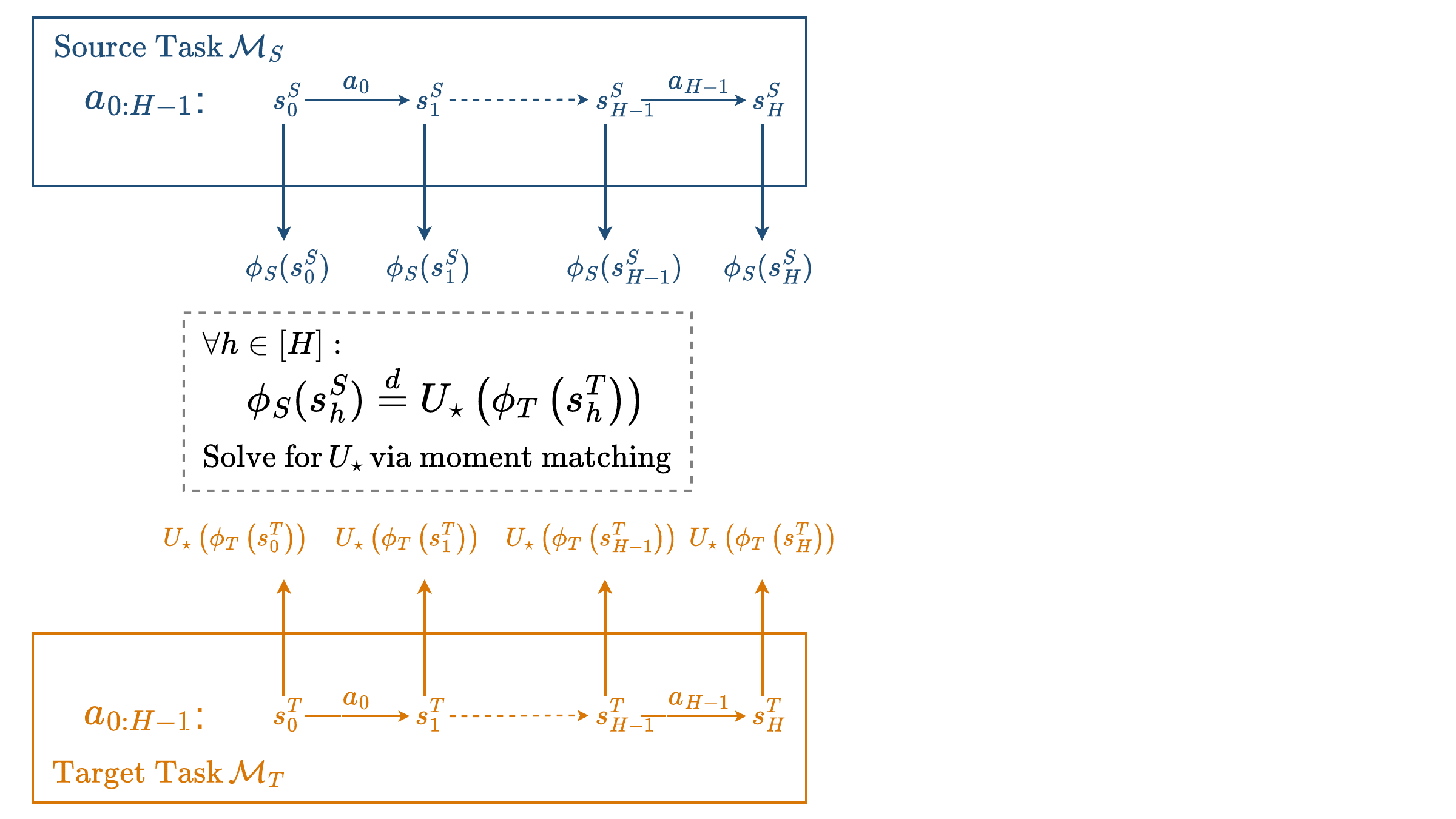}
    \caption{Overview of {\alggen}. We match state feature statistics between source and \emph{undone} target trajectories.}
    \label{fig:overview}
    \vskip -0.6cm
\end{wrapfigure}

In this section, we show how the undo map can be learned by leveraging the invariance of the action space across the source and target MDPs. First, we formally define a distributional measure to evaluate the quality of a candidate undo map (Section~\ref{sec:dist_undo}). We then propose a practical learning algorithm, {\alggen} (Section~\ref{sec:learn_undo}), that matches observation statistics between the source MDP and the target MDP viewed through an undo map. This algorithm, illustrated in Fig.~\ref{fig:overview}, enables transfer of the source policy by composing it with the learned undo map. Theoretically, under certain assumptions, we show that {\alggen} enjoys strictly better sample complexity than learning from scratch (Section~\ref{sec:theory}).

\subsection{Distributional Measure of Undo Map Quality}
\label{sec:dist_undo}

To estimate the undo map, we match state feature distributions induced by an \emph{open-loop} sequence of actions in the source MDP $\mathcal{M}_S$ with those induced by the \emph{undone} trajectories in the target MDP $\mathcal{M}_T$. This leverages the fact that while the observations differ, the underlying dynamics remain consistent.

For an arbitrary undo map $U$, we now introduce the notion of source and \emph{undone} target state features.

\begin{definition}[Source and Undone Target Features]
Given an open-loop action sequence $a_{0:h-1}$, define the source feature random variable at timestep $h$ as
\begin{equation}
\psi_h^{S}(a_{0:h-1}) \triangleq \phi_S(s_h),
\end{equation}
and, for an arbitrary undo map $U$, define the \emph{undone} target feature random variable as
\begin{equation}
\psi_{h,U}^{T}(a_{0:h-1}) \triangleq U\!\left(\phi_T(s_h)\right),
\end{equation}
where $(s_0,\ldots,s_h)$ is the state sequence obtained by executing $a_0,\ldots,a_{h-1}$ in the respective MDP.
\end{definition}

\begin{algorithm*}[t]
\caption{Learning to Undo with {\alggen}}
\label{alg:undo-map-gen}
\begin{algorithmic}[1]
\State \textbf{Input:} source $\mathcal{M}_S$, target $\mathcal{M}_T$, trajectories $n$, iterations $N$, moments $M$, truncated horizon $H^\dagger$
\State Initialize undo map parameters $\phi$.

\For{$i \in [N]$} 
    \State Sample an action sequence $a_{0:H^\dagger-1}$ uniformly at random.
    \Statex \ \ \quad \LINECOMMENT{Exactly compute source feature mean and marginal central moments.}
    \State Compute mean $\mu_S(h) \leftarrow \mathbb{E}[\psi_h^{S}(a_{0:h-1})]$, $\forall h \in [H^\dagger]$.
    \State Compute moments $c_{k,S}(h) \leftarrow \mathbb{E}\big[(\psi_h^{S}(a_{0:h-1})-\mu_S(h))^{\odot k}\big]$, $\forall k \in [M]$, $\forall h \in [H^\dagger]$.
    
    \Statex \ \ \quad \LINECOMMENT{Estimate \emph{undone} target feature mean and marginal central moments.}
    \State Collect $n$ trajectories in $\mathcal{M}_T$ using action sequence $a_{0:H^\dagger-1}$.
    \For{$h \in [H^\dagger]$}
        \State For each trajectory $j$, compute \emph{undone} target features $\psi_{h, U_\phi}^{T, (j)}(a_{0:h-1})$.
        
        \State Compute mean $\hat{\mu}_T(h) \gets \frac{1}{n}\sum_{j=1}^n \psi_{h, U_\phi}^{T, (j)}(a_{0:h-1})$.
        \State Compute moments $\hat{c}_{k,T}(h) \gets \frac{1}{n}\sum_{j=1}^n \big(\psi_{h, U_\phi}^{T, (j)}(a_{0:h-1})-\hat{\mu}_T(h)\big)^{\odot k}$, $\forall k \in [M]$.
    \EndFor
    
    \Statex \ \ \quad \LINECOMMENT{Minimize empirical undo discrepancy.}
    \State Update $\phi$ to minimize $\sum_{h=0}^{H^\dagger-1} \Big[\|\mu_S(h)-\hat{\mu}_T(h)\|_2 + \sum_{k=2}^M \|c_{k,S}(h)-\hat{c}_{k,T}(h)\|_2\Big]$.
\EndFor
\State \textbf{Output:} undo map parameters $\phi$
\end{algorithmic}
\end{algorithm*}

By the definition of the true undo map $U_\star$, we have for all $h \in [H]$,
\begin{equation}
\psi_h^{S}(a_{0:h-1}) \;\overset{d}{=}\; \psi_{h,U_\star}^{T}(a_{0:h-1}),
\label{eq:undo_rel_gen}
\end{equation}
where $\overset{d}{=}$ denotes equality in distribution, which means that applying $U_\star$ to the target observations recovers exactly the source distribution, conditioned on any open-loop action sequence $a_{0:h-1}$.

Motivated by Eq.~\ref{eq:undo_rel_gen}, our goal is to evaluate a candidate undo map $U$ by measuring how closely the \emph{undone} target features match the source features in distribution, under open-loop rollouts. To quantify distributional mismatch, we use the Central Moment Discrepancy (CMD) metric \citep{zellinger2017cmd}, which compares distributions via their means and central moments.

\begin{definition}[Central Moment Discrepancy (CMD) \citep{zellinger2017cmd}]
Let $p$ and $q$ be probability distributions supported on a compact domain $[a,b]^N$, and let $X\sim p$ and $Y\sim q$ be bounded random vectors. The Central Moment Discrepancy (CMD) between $p$ and $q$ is defined as
\begin{equation}
\begin{aligned}
\mathrm{CMD}(p,q) \triangleq\; &\frac{1}{|b-a|} \big\|\mathbb{E}[X]-\mathbb{E}[Y]\big\|_2 + \sum_{k=2}^{\infty} \frac{1}{|b-a|^k} \big\|c_k(X)-c_k(Y)\big\|_2 ,
\end{aligned}
\label{eq:cmd}
\end{equation}
where $c_k(\cdot)$ denotes the $k$-th order central moment vector.
\end{definition}

$\mathrm{CMD}(p,q)=0$ iff $p=q$. Moreover, if $\mathrm{CMD}(p_n,p)\to 0$, then $p_n \xrightarrow{d} p$, where $\xrightarrow{d}$ denotes convergence in distribution \citep{zellinger2017cmd}. 

For notational convenience, given random variables $X$ and $Y$, we write $d_{\mathrm{CM}}(X,Y)$ to denote the CMD between the distributions induced by $X$ and $Y$. We now use this to define a measure for the quality of an arbitrary undo map.

\begin{definition}[Undo Discrepancy]
For an undo map $U$, we define its discrepancy as
\begin{equation}
\mathcal{D}(U)
\triangleq
\mathbb{E}_{a_{0:H-1}}
\Bigg[
\sum_{h=0}^{H-1}
d_{\mathrm{CM}}\Big(
\psi_h^{S}(a_{0:h-1}),
\psi_{h,U}^{T}(a_{0:h-1})
\Big)
\Bigg],
\label{eq:undo-discrepancy}
\end{equation}
where $\psi_h^{S}(a_{0:h-1})$ and $\psi_{h,U}^{T}(a_{0:h-1})$ denote the source and
\emph{undone} target feature random variables at timestep $h$, and the expectation is over uniformly sampled action sequences.
\end{definition}

Thus, minimizing $\mathcal{D}(U)$ is equivalent to matching the \emph{undone} feature distributions in the target to the corresponding feature distributions in the source at each timestep for open-loop sequences of actions.

\begin{figure*}[t]
    \centering
    \includegraphics[
        width=0.88\textwidth
    ]{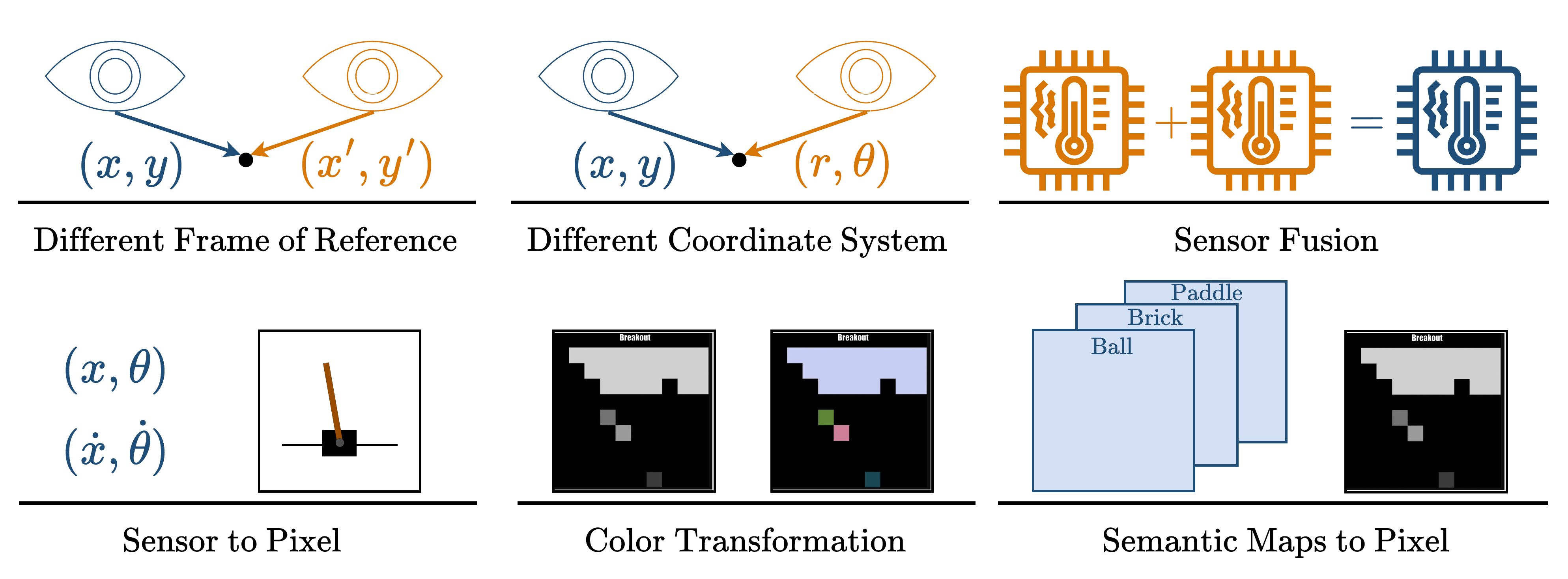}
    \caption{We evaluate {\alggen} under structured transformations ranging from geometric and sensor-level changes to gaps in the level of abstraction (e.g., semantic maps in the source and pixels in the target), where the source task is often easier to learn than the target.}
    \label{fig:env_vis}
\end{figure*}

\subsection{Learning the Undo Map}
\label{sec:learn_undo}

With a distributional measure of undo map quality defined, we now describe how to learn the map.

\textbf{Empirical Undo Discrepancy.}
To empirically estimate $\mathcal{D}(U)$ from samples, we approximate CMD using only the first $M$ moments and compute empirical marginal central moments across state feature coordinates. Concretely, given samples $\{x_i\}_{i=1}^n \sim p$, the $k$-th order empirical marginal central moment vector, $\hat{c}_k(p)$, is given by $\frac{1}{n}\sum_{i=1}^n (x_i-\hat{\mu}_p)^{\odot k}$, where $(\cdot)^{\odot k}$ denotes element-wise exponentiation and $\hat{\mu}_p = \frac{1}{n}\sum_{i=1}^n x_i$ is the empirical mean.

\textbf{Learning Algorithm.}
We parameterize the undo map as $U_\phi$ and optimize its parameters $\phi$ to match state feature distributions in the source and the \emph{undone} target. More concretely, we minimize the empirical undo discrepancy $\widehat{\mathcal{D}}(U_\phi)$, defined via a truncated empirical CMD estimator using the first $M$ marginal central moments. Our algorithm samples multiple action sequences and optimizes these moments over the resulting rollouts. As samples from the source MDP $\mathcal{M}_S$ are free, we can compute these moments in the source exactly, while they are empirically estimated using samples from the target MDP $\mathcal{M}_T$. Moreover, we use a truncated horizon $H^\dagger < H$ to allow more action sequences to be sampled for training under a given budget of samples; we justify this choice in Section~\ref{sec:experiments}.

Algorithm~\ref{alg:undo-map-gen} gives the pseudocode for our algorithm, {\alggen}. In summary, we iteratively sample an action sequence, compute marginal central moments of source state features, estimate marginal central moments of the corresponding \emph{undone} target state features via $n$ trajectories, and update $\phi$ to minimize the empirical discrepancy. In practice, the source expectations are also estimated from samples. We then transfer the source policy to the target by composing it with the learned undo map.

\subsection{Theoretical Analysis}
\label{sec:theory}

In this section, we show that {\alggen} is more sample efficient than learning from scratch. For this analysis, we specialize our formalism to state-action features used to represent the optimal $Q$-function. We assume that the source and target state-action features are related by an undo map. Recovering it allows the source optimal $Q$-function (and therefore its greedy policy) to be transferred to the target.

We first list the assumptions used in our analysis. The source and target MDPs have feature maps $\phi_S: \mathcal{S}^S \!\times\! \mathcal{A} \to \mathbb{R}^{d_S}$ and $\phi_T: \mathcal{S}^T \!\times\! \mathcal{A} \to \mathbb{R}^{d_T}$ that represent state–action pairs as $d_S$ and $d_T$-dimensional vectors. Let $\Phi_S \in \mathbb{R}^{|\mathcal{S}^S \!\times\! \mathcal{A}| \times d_S}$ and $\Phi_T \in \mathbb{R}^{|\mathcal{S}^T \!\times\! \mathcal{A}| \times d_T}$ denote the corresponding feature matrices, with rows $\phi_S(s,a)$ and $\phi_T(s,a)$. We assume that the feature vectors in both MDPs satisfy $\|\phi(s, a)\|_2 \le 1$, and that rewards are bounded within the range $[-1, 1]$. We further assume that the source is linearly-$Q^\star$ realizable.

\begin{assumption}[Linear-$Q^\star$ realizability in $\mathcal{M}_S$ \citep{pmlr-v132-weisz21a}]
\label{assump:realizability}
There exists a vector $\theta^\star_S \in \mathbb{R}^{d_S}$ such that for every state–action pair $(s,a) \in \mathcal{S}^S \times \mathcal{A}$ in the source MDP $\mathcal{M}_S$,
\[
    Q^\star_S(s,a) \;=\; \langle \phi_S(s,a), \theta^\star_S \rangle.
\]
\end{assumption}

In this setup, given an action sequence $a_{0:h-1}$, we define $\psi_h(a_{0:h-1})$ to be the state-action feature random variable at timestep $h-1$:
\begin{equation}
\psi_h(a_{0:h-1}) \triangleq \phi(s_{h-1}, a_{h-1}).
\end{equation}

The source and the target are exactly the same except for the state-action features, which are related via a linear undo map $U_\star$: $\Phi_S = \Phi_T U_\star^T$. We assume that the $\ell_2$ norm of each row of $U_\star$ is upper bounded by a constant. In the source, the learner has access to the parameter vector $\theta^\star_S$ for the state-action values of the optimal policy, and an oracle $\mathcal{O}$ for the expected state-action features.

\begin{assumption}[Oracle for expected state-action features in $\mathcal{M}_S$]
    We denote by $\mathcal{O}$ an oracle that when queried with $a_{0:h-1}$ and $h \in [H]$, outputs $\bar{\psi}_h^S(a_{0:h-1}) \triangleq \mathbb{E}\left[ \psi^S_h(a_{0:h-1}) \right]$.
\end{assumption}

We would like to note that this assumption is only for convenience; the same analysis holds when the source expectations are estimated empirically via random-design linear regression arguments.

Since the source and target MDPs differ only through a linear transformation of the state-action features, the expected state-action features in the source and the target (denoted by $\bar{\psi}^S_h(a_{0:h-1})$ and $\bar{\psi}^T_h(a_{0:h-1})$, respectively) are related via the undo map $U_\star$ as follows: 

\begin{equation}
\bar{\psi}_h^S(a_{0:h-1}) = U_\star \bar{\psi}_h^T(a_{0:h-1}),\, \forall h \in [H].
\label{eq:undo_rel}
\end{equation}

\begin{figure*}[t]
    \centering
    \includegraphics[width=\textwidth]{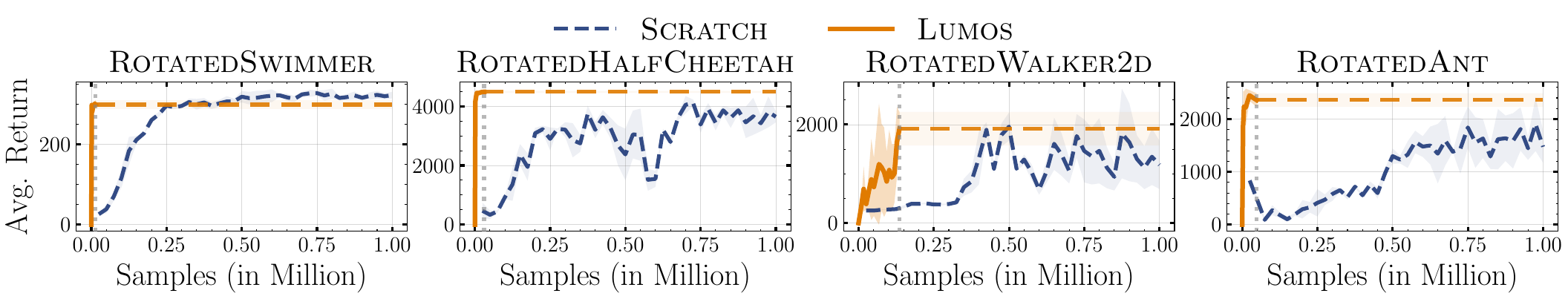} \\
    \includegraphics[width=\textwidth]{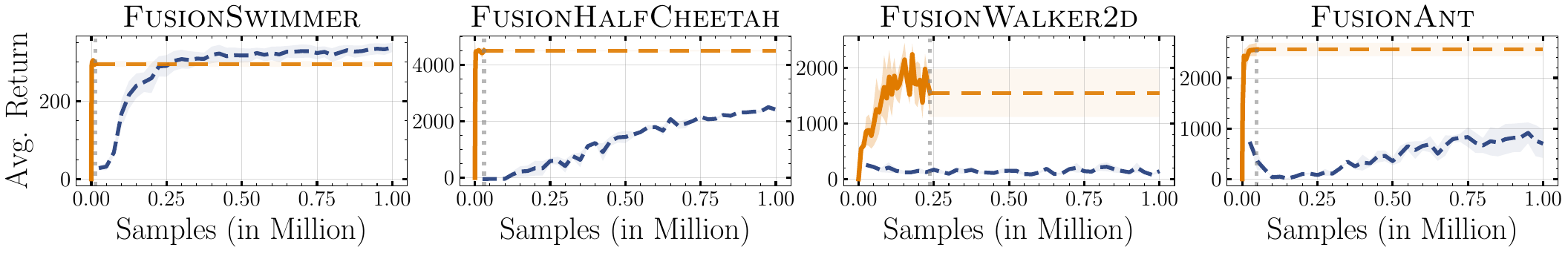}
    \caption{Comparison of avg. return (standard deviation over five random seeds) for varying budgets of target samples. Transfer learning via {\alggen} consistently achieves significantly better sample efficiency than learning from scratch.}
    \label{fig:main_result_lin}
    \vspace{-0.5cm}
\end{figure*}

\textbf{Remark.}
When the undo map is linear, higher-order central moments of the transformed features are completely determined by the first-order statistics through the same linear operator. Therefore, matching higher-order moments provides no additional constraints beyond mean matching.

\textbf{Learning Algorithm.}
Our learning algorithm first queries $\mathcal{O}$ to obtain $\bar{\psi}_h^S(a_{0:h-1})$ in $\mathcal{M}_S$, for all $h \in [H]$. Next, it rolls out $n$ trajectories in the target using the open-loop action sequence $a_{0:H-1}$. It then views the state-action features $\psi_h^{T, (i)}(a_{0:h-1})$ (from the $i$-th sample trajectory) as noisy features for $\bar{\psi}_h^S(a_{0:h-1})$. Subsequently, it estimates $U_\star$ with $\widehat{U}$ via least squares.

\textbf{Policy Transfer.} From Eq.~\ref{eq:undo_rel}, the relation $Q^\star_T = \Phi_T U_\star^T \theta^\star_S$ holds. Thus, a natural way to evaluate the output $\widehat{U}$ of {\alggen} is to bound the performance difference with respect to the optimal policy in the target, if we act greedily with respect to $\Phi_T \widehat{U}^T \theta^\star_S$. Let \(\hat{\pi}\) denote this policy. 

We show that our algorithm has the following guarantee; the proof is provided in Appendix~\ref{sec:app.proofs}.

\begin{theorem}[Performance Difference Bound for {\alggen}]
Define the covariance matrix
\[
    \Sigma \triangleq \frac{1}{H} \sum_{h=1}^H \bar{\psi}_h^{T}(a_{0:h-1}) \bar{\psi}_h^{T}(a_{0:h-1})^T,
\]
and let $\lambda_{\min}$ denote $\lambda_{\min}(\Sigma)$.

For all
\(n \geq \frac{6}{H \lambda_{\min}} (\log \frac{3 d_T d_S}{\delta})\), with probability at least \(1 - \delta\), the performance difference satisfies
\[
\begin{aligned}
\|V_T^\star - V_T^{\hat{\pi}}\|_\infty
&\leq
\widetilde{O}\left(\sqrt{\frac{H}{n\lambda_{\min}}}
\sqrt{
d_T
+ 2\log\!\left( \frac{3d_S}{\delta} \right)}\right).
\end{aligned}
\]

\label{thm:perf-bound}
\end{theorem}

The theorem shows that if the \emph{open-loop} action sequence $a_{0:H-1}$ is \emph{good}, i.e., $\lambda_{\min}(\Sigma)$ is sufficiently large (on the order of $1/d_T$), then {\alggen} achieves a sample complexity upper bound that scales as $\widetilde{O}(d_T\sqrt{H})$. This improves upon the lower bound for linearly-$Q^\star$ realizable MDPs, which scales exponentially with the feature dimension $d_T$ or the horizon $H$ \citep{pmlr-v132-weisz21a}.

\textbf{Remark on Linear MDPs.} A special case of MDPs with linear-$Q^\star$ realizability is linear MDPs \citep{pmlr-v125-jin20a}. In this setting, the minimax lower bound has a $d_T H^{3/2}$ dependence on $d_T$ and $H$ \citep{pmlrhe23d}, whereas {\alggen} enjoys a $d_T \sqrt{H}$ dependence, up to logarithmic factors.


\section{Experimental Evaluation}
\label{sec:experiments}

In this section, we empirically evaluate our method to address the following research questions: 
\begin{enumerate*}[label=(\roman*)]
    \item Does transfer via {\alggen} outperform learning from scratch? Concretely, does learning the undo map $U_\star$ yield better target policies than single-task training?
    \item How does {\alggen} compare with transfer learning baselines that assume reward feedback in the target, which is an orthogonal setting to ours?
    \item Does matching higher-order moments contribute to performance gains?
    \item How should the training budget be allocated between the number of iterations $N$ and the truncated horizon $H^\dagger$?
\end{enumerate*}

We begin by describing the transfer settings. Next, we provide an overview of the training process, followed by empirical results. We additionally analyze the robustness of {\alggen} in Appendix~\ref{sec:app.robustness}.

\begin{figure*}[t]
    \centering
    \includegraphics[width=0.7\textwidth]{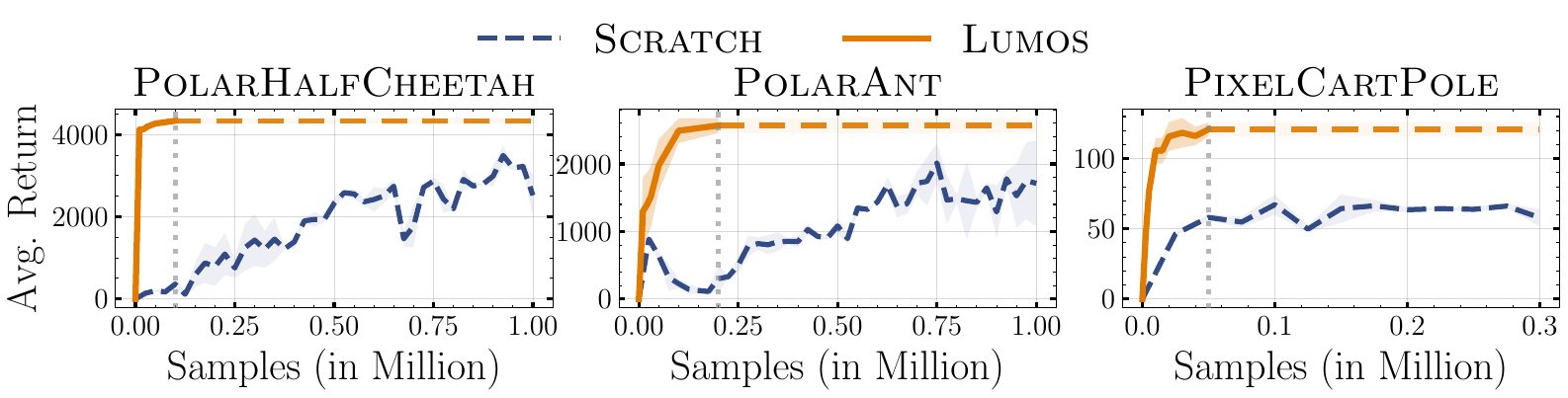} \\
    \includegraphics[width=0.7\textwidth]{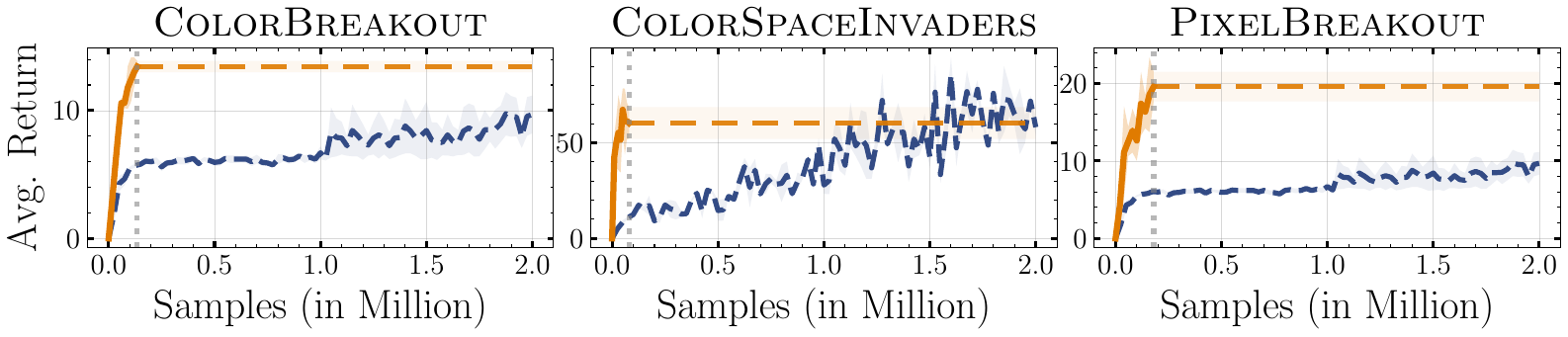}
    \caption{Comparison of avg. return (standard deviation over five random seeds) for varying budgets of target samples. Transfer learning via {\alggen} consistently achieves significantly better sample efficiency than learning from scratch.}
    \label{fig:main_result_nonlin}
    \vspace{-0.5cm}
\end{figure*}

\begin{figure*}[t]
    \centering
    \includegraphics[width=\textwidth]{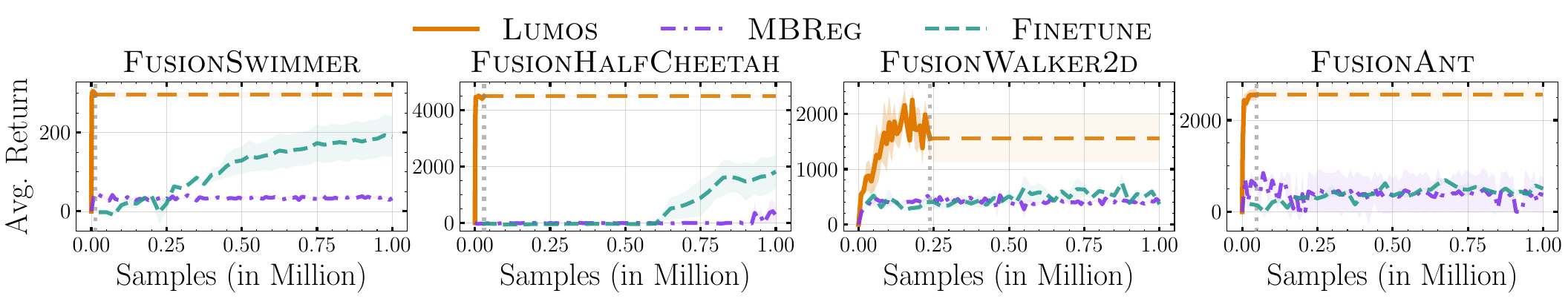}
    \caption{Comparison of avg. return (standard deviation over five random seeds). {\alggen} consistently achieves better sample efficiency than transfer baselines that require rewards in the target.}
    \label{fig:main_result_baseline}
    \vspace{-0.5cm}
\end{figure*}

\subsection{Environments}

We evaluate our method on a suite of continuous control tasks and Atari games that are challenging to learn from scratch. As source tasks, we consider standard MuJoCo environments: \textsc{Swimmer}, \textsc{HalfCheetah}, \textsc{Walker2d}, \textsc{Ant}, and \textsc{CartPole}, as well as Atari games: \textsc{Breakout} and \textsc{SpaceInvaders}. We construct corresponding target tasks by applying structured transformations to the state space. Detailed descriptions are provided in Appendix~\ref{sec:app.env_desc}.

\subsection{Training Procedure}
We use the \texttt{rl-baselines3-zoo} framework \citep{rl-zoo3} to train both the source policies and the target policies used in the learning-from-scratch baseline, with PPO \citep{PPO} as the learning algorithm. For each environment, we adopt the tuned hyperparameters provided in the framework. Our Atari experiments are based on the \texttt{MinAtar} benchmark \citep{minatar}. 

To estimate the state feature means and marginal central moments for undo discrepancy, we use a large number of samples from the source MDP to obtain accurate statistics, while using fewer samples from the target. We employ multiple short action sequences for good coverage of the features. 

The undo map is parameterized by a neural network and trained via {\alggen}. Additional implementation details are provided in Appendix~\ref{sec:app.impl_details}. 

\vspace{-0.2cm}
\subsection{Results}
In this section, we first compare our method, {\alggen}, against learning the target policy from scratch (listed as \textsc{Scratch}). Next, we evaluate the contribution of higher-order moments to performance gains. Finally, we examine the effect of using a truncated horizon $H^\dagger$.

\textbf{Results with {\alggen}.}
We evaluate {\alggen} in transfer settings that go beyond the regime of our theoretical analysis. Figures~\ref{fig:main_result_lin} and~\ref{fig:main_result_nonlin} report the average return (with standard deviation over five random seeds) as a function of the target sampling budget. We stop \textsc{Lumos} early upon reaching near-source performance, after which the dashed line extends the final performance to larger budgets.

Across all tasks, {\alggen} achieves consistent and substantial improvements over single-task training (\textsc{Scratch}). The \textsc{Fusion} and \textsc{Pixel} targets are especially challenging for learning from scratch, due to the increased observation dimensionality and the fact that useful information must be recovered through structured aggregation (across sensors in \textsc{Fusion}, or from high-dimensional visual inputs in \textsc{Pixel}). In contrast, {\alggen} explicitly exploits this shared structure between source and target to achieve near-optimal performance with few target samples.

Overall, these results provide empirical evidence that undo-map-based transfer leads to sample-efficiency gains in RL under structured state-space transformations.

\textbf{Comparison with transfer RL baselines.} While our setup assumes no reward feedback in the target, for completeness we compare {\alggen} against transfer RL baselines that operate with reward feedback in the target. Figure~\ref{fig:main_result_baseline} includes a standard transfer baseline (\textsc{Finetune}) that initializes the target policy from the source policy (except for the input layer when dimensions differ). We also evaluate \textsc{MBReg} \citep{lit_tl_1}, a baseline that learns a reward and dynamics model over latent states in the source and uses it for model regularization in the target. Despite the favorable conditions afforded to these baselines (namely, access to reward feedback in the target), {\alggen} consistently outperforms them across a variety of environments.

\begin{figure*}[t]
    \centering
    \includegraphics[height=2.9cm,keepaspectratio]{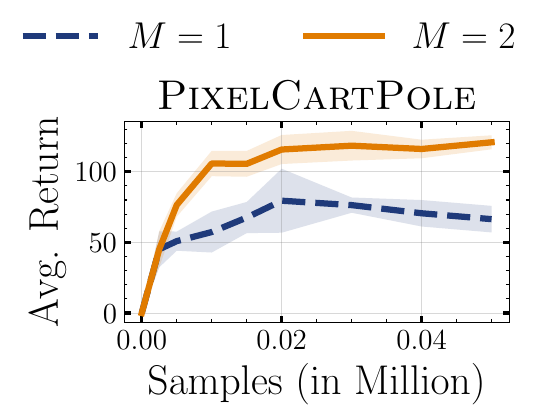}
    \includegraphics[height=3.3cm,keepaspectratio]{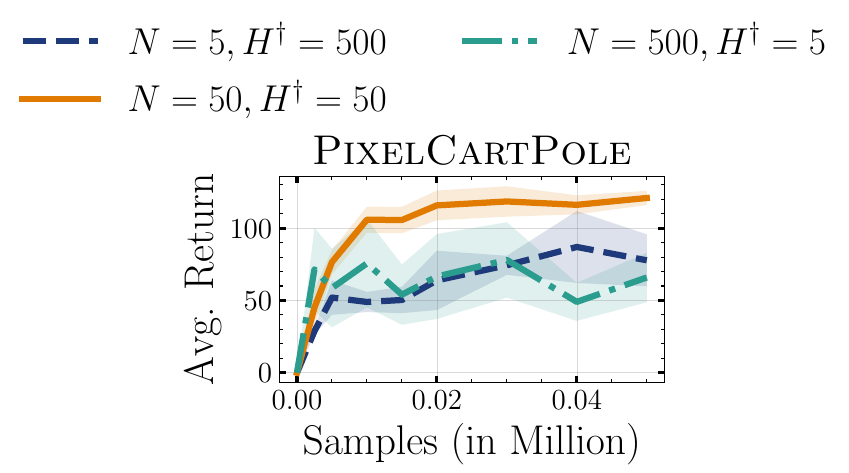}
    \caption{Ablation results. \textbf{(a)} Effect of the number of matched moments $M$. \textbf{(b)} Effect of allocating a fixed sampling budget between the number of action sequences $N$ and their truncated horizon $H^\dagger$.}
    \label{fig:ablations}
    \vspace{-0.5cm}
\end{figure*}

\textbf{Ablation w.r.t.\ the number of moments.}
We use \textsc{PixelCartPole} to study the effect of matching higher-order central moments when learning the undo map. Specifically, we run {\alggen} with $M=1$ (mean matching only) and $M=2$ (including second-order moments). Figure~\ref{fig:ablations}(a) shows that using higher-order moments yields clear performance improvements.

\textbf{Ablation w.r.t.\ the truncated horizon.}
We use \textsc{PixelCartPole} to justify our algorithmic choice of truncating trajectories at horizon $H^\dagger$. Keeping the overall sampling budget fixed, we vary the allocation between the number of sampled action sequences $N$ and their length $H^\dagger$. Concretely, we consider the configurations $(N, H^\dagger) \in \{(5,500), (50,50), (500,5)\}$, with the product $N H^\dagger$ kept constant. Figure~\ref{fig:ablations}(b) indicates that moderately short sequences perform best: using shorter sequences improves performance relative to very long rollouts (e.g., $(50,50)$ vs.\ $(5,500)$), while overly short horizons can hurt due to potentially insufficient state-space coverage (e.g., $(500,5)$).


\vspace{-0.3cm}
\section{Related Work}
\vspace{-0.1cm}

\textbf{Transfer Learning.}
Empirical methods for transfer learning in RL often lack provable guarantees or rely on application-specific assumptions that do not generalize. For instance, \citet{lit_tl_1} assume shared latent dynamics across observation spaces and that target observations can be transformed into the source domain; they do not learn this mapping explicitly and offer no theoretical guarantees on transfer performance. Similarly, \citet{lit_tl_5} assume a shared latent MDP structure between tasks but their method is not theoretically grounded. Other works, such as \citet{lit_tl_2}, study transfer through reusing prior policies across design iterations, which is orthogonal to our setting. Methods like \citet{lit_tl_3, lit_tl_4} rely on very specific visual or object-level assumptions, which limits their general applicability. The notion of an \emph{undo} map to reverse state space transformations in reinforcement learning was first introduced in \citet{undo}, where the problem is approached through a distributional lens by aligning trajectories across the source and target tasks. However, the method does not scale beyond toy tabular environments. Orthogonal to our setup, works such as \citet{lit_add_1, lit_add_2} study transfer under dynamics shift.

\textbf{Representational Transfer in Low-Rank MDPs.} 
Several works have studied representational transfer under low-rank or linear MDP assumptions. For instance, \citet{lit_rl_1, lit_rl_3} use reward-free exploration in the source task to learn a good representation for the target. \citet{lit_rl_2} extend this to a more general low-rank setting. \citet{lit_rl_4} learn a linear representation using least squares for multitask linear MDPs, and \citet{lit_rl_5} consider the offline multitask case. \citet{composite} provide guarantees for transfer Q-learning in the composite MDP setting, but their method does not scale empirically and only works for finite state-action spaces.

In contrast, our work approaches transfer RL through a more principled lens by making concrete assumptions about the structural relationship between the source and target tasks. We propose an algorithm that explicitly learns a mapping between their observation spaces by framing the problem as supervised learning, and therefore eliminates the need for RL in the target task. Under certain assumptions, this leads to provable improvements in sample complexity over learning from scratch. Moreover, our method is practical and demonstrates strong empirical performance. Essentially, our work takes a step toward more principled and broadly applicable approaches to transfer in RL.

Our goal is not to develop a state-of-the-art transfer RL method competing across benchmarks, but rather to design a theoretically grounded approach that is provably better than learning from scratch when suitable shared structure exists, and is also empirically better than single-task RL.
\vspace{-0.3cm}
\section{Concluding Discussion}
\vspace{-0.2cm}
In this paper, we introduced a principled approach to transfer learning in reinforcement learning, where the source and target tasks are related by structured transformations of the state space. By explicitly learning an undo map, our method achieves significant gains in sample efficiency compared to learning target policies from scratch, as demonstrated across a range of challenging continuous control tasks and Atari games. Theoretically, we showed that when the undo map is linear and the source is linearly-$Q^\star$ realizable, our approach achieves strictly better sample complexity than learning from scratch. A promising direction for future work is to study more general notions of task equivalence, such as MDP homomorphisms where the action space may also be transformed. Extending our guarantees beyond the linear realizability assumption is another interesting direction.

\textbf{Broader impact.} Our work is algorithmic, and we do not see any direct societal implications.

\textbf{Funding.} No external funding was received.

\textbf{Competing Interests.} The authors declare no competing interests.
\bibliographystyle{plainnat}
\bibliography{main}
\appendix
\onecolumn
{
    \allowdisplaybreaks
\section{Table of Contents}
\label{sec:app.toc}

In this section, we provide an outline of the contents provided in the paper's appendices.
\begin{itemize}[leftmargin=*]
    \item Appendix~\ref{sec:app.proofs} contains the proof of Theorem~\ref{thm:perf-bound}.
    \item Appendix~\ref{sec:app.env_desc} describes the environments used in the experimental evaluation.
    \item Appendix~\ref{sec:app.impl_details} provides additional implementation details for the experimental evaluation.
    \item Appendix~\ref{sec:app.robustness} provides additional results examining the robustness of {\alggen}.
\end{itemize}

\section{Proofs}\label{sec:app.proofs}
In this section, we provide the proof of Theorem~\ref{thm:perf-bound}.

\textbf{Proof Structure.} The overall structure of our proof is:
\begin{enumerate}
    \item View the estimation of each row of $\widehat U$ as a random-design linear regression problem and bound its covariance-weighted estimation error.
    \item Use $\lambda_{\min}(\Sigma)$ to convert this covariance-weighted error into a max-norm bound on $\widehat U-U_\star$.
    \item Use the relation $Q_T^\star=\Phi_TU_\star^\top\theta_S^\star$ to translate undo-map estimation error into error in the approximate target $Q$-function $\widehat{Q_T} = \Phi_T\widehat U^\top\theta_S^\star.$
    \item Apply a standard performance bound for the policy greedy with respect to $\widehat{Q_T}$ to obtain the final value-difference guarantee.
\end{enumerate}

We first relate the performance difference between the optimal policy in the target $\mathcal{M}_T$ and the policy that acts greedily with respect to $\Phi_T \widehat{U}^T \theta^\star_S$ to the estimation error in $\widehat{U}$, where $\widehat{U}$ estimates $U_\star$.

\begin{lemma}[Performance Difference]
$\|V_T^\star - V_T^{\hat{\pi}}\|_\infty \leq 2H \|\widehat{U} - U_\star\|_{\max}$.
\label{lem:perf_diff}
\end{lemma}

Here, $\|\cdot\|_{\max}$ denotes the max norm. To bound the estimation error $\|\widehat{U} - U_\star\|_{\max}$, we use the following per-row guarantee:

\begin{lemma}[Estimation Error Bound]  
Let \( M[i] \) denote the $i$-th row of a matrix $M$. Fix \( i \in [d_S] \). Then, for all \(n \geq \frac{6}{H\lambda_{\min}} (\log \frac{3 d_T}{\delta})\), with probability at least \( 1 - \delta \), the output $\widehat{U}$ of {\alggen} satisfies:
\[
    \norm{\widehat{U}[i] - U_\star[i]}_{\Sigma} \leq \sqrt{\frac{1}{nH}} \sqrt{ d_T + 2\log\left( \frac{3}{\delta} \right) + 2 \sqrt{d_T \log\left( \frac{3}{\delta} \right)}} + o\left(\sqrt{\frac{1}{nH}}\right).
\]
\label{lem:rr_guarantee}
\end{lemma}

Since \( \Sigma \) is positive semidefinite (PSD), we can relate the $\Sigma$-norm to the $\ell_\infty$ norm via:
\begin{equation}
    \norm{x}_\infty \leq \frac{\norm{x}_{\Sigma}}{\sqrt{\lambda_{\min}}}.
\end{equation}

Applying a union bound over all \( i \in [d_S] \), with per-row failure probability set to \( \delta/d_S \), gives a high-probability bound on \( \|\widehat{U} - U_\star\|_{\max} \). Substituting this into Lemma~\ref{lem:perf_diff} concludes the proof.

\textbf{Proof of Lemma~\ref{lem:perf_diff}.}
Let $f$ denote the approximate target $Q$-function induced by the estimated undo map: $f(s,a) \triangleq \left\langle \widehat U \phi_T(s,a), \theta_S^\star \right\rangle$. For any $s \in \mathcal{S}^T$,
\begin{align*}
V_T^\star(s) - V_T^{\hat{\pi}}(s)
&= Q_T^\star(s, \pi^\star(s)) - Q_T^\star(s, \hat{\pi}(s)) + Q_T^\star(s, \hat{\pi}(s)) - Q_T^{\hat{\pi}}(s, \hat{\pi}(s)) \\
&\leq Q_T^\star(s, \pi^\star(s)) - f(s, \pi^\star(s)) + f(s, \hat{\pi}(s)) - Q_T^\star(s, \hat{\pi}(s)) \\
&\quad + \mathbb{E}_{s' \sim \mathbb{P}^T(s, \hat{\pi}(s))}\left[ V_T^\star(s') - V_T^{\hat{\pi}}(s') \right] \\
&\leq 2 \|f - Q_T^\star\|_\infty + \mathbb{E}_{s' \sim \mathbb{P}^T(s, \hat{\pi}(s))}\left[ V_T^\star(s') - V_T^{\hat{\pi}}(s') \right].
\end{align*}

Unrolling this recursion for \( H \) steps,
\[
\|V_T^\star - V_T^{\hat{\pi}}\|_\infty \leq 2H \|f - Q_T^\star\|_\infty.
\]

Substituting \( f = \Phi_T \widehat{U}^\top \theta^\star_S \) and \( Q_T^\star = \Phi_T U_\star^\top \theta^\star_S \) completes the proof.

\textbf{Proof of Lemma~\ref{lem:rr_guarantee}.}
Fix a source-feature coordinate $i\in[d_S]$. Estimating $U_\star[i]$ can be viewed as a random-design linear regression problem in which the target state-action features (of dimensionality $d_T$) are the covariates and the corresponding expected source-feature coordinate is the response. The regression parameter is $U_\star[i]$, and the covariance matrix is $\Sigma$. \citet{random_ridge} bound the covariance-weighted estimation error of least squares under three conditions.

In our setting:
\begin{enumerate}
    \item The regression noise is bounded in $[-1,1]$ and is therefore $1$-subgaussian, so we set $\sigma=1$.
    \item The linear model is correctly specified through $\overline\psi_h^S = U_\star\overline\psi_h^T$. Therefore, there is no approximation bias, and we set $b_0=0$.
    \item Since the $\ell_2$-norm of $\phi_T(s,a)$ is at most $1$ and the minimum eigenvalue of $\Sigma$ is $\lambda_{\min}$, the statistical leverage parameter $\rho_0$ satisfies $\rho_0 \leq \frac{1}{\sqrt{\lambda_{\min}d_T}}$ by Remark 2 of \citep{random_ridge}.
\end{enumerate}

Substituting these quantities into Theorem~1 and Remark~9 of \citep{random_ridge} completes the proof.

\section{Environments}\label{sec:app.env_desc}

\subsection{Types of Transformations}
\textbf{Changed frame of reference.}
Here, we rotate the global coordinate frame in the source tasks, which results in rotated variants of the MuJoCo tasks (e.g., \textsc{RotatedAnt}). This models geometric discrepancies that arise, for example, from changes in coordinate conventions across simulators.

\textbf{Sensor fusion.}
In this setting, multiple sensor modalities are available, which results in sensor-fused variants (e.g., \textsc{FusionAnt}). Instead of observing a $d_S$-dimensional vector, the agent receives a $10d_S$-dimensional observation ($10\times$ sensors), which can be combined to recover the original values. This captures settings where sensors differ at deployment.

\textbf{Changed coordinate system.}
We construct polar-coordinate variants \textsc{PolarHalfCheetah} and \textsc{PolarAnt}, where sensor readings are expressed in polar rather than Cartesian coordinates. This captures settings with differing state parameterizations.

\textbf{Sensor-to-pixel observations.}
We consider \textsc{CartPole} with sensor observations as the source task and pixel-based observations as the target, which results in \textsc{PixelCartPole}.
This introduces an abstraction gap where the source provides structured low-dimensional inputs, while the target requires learning directly from raw visual observations.

\textbf{Color transformation.}
For Atari games, we treat grayscale observations as the source and RGB observations as the target, which results in the variants \textsc{ColorBreakout} and \textsc{ColorSpaceInvaders}. This setting models rendering shifts at deployment time.

\textbf{Semantic bitmaps to pixels.}
Finally, we consider a setting where the source provides object-centric semantic bitmaps, but the target provides raw pixel observations (\textsc{PixelBreakout}). This represents an abstraction gap where learning from scratch is significantly harder.

\subsection{Environment-Specific Details}

\textbf{\textsc{RotatedSwimmer}.}
This task requires the agent to swim forward through a viscous fluid. We apply a $45^\circ$ rotation to the $(x, y)$ plane to modify the frame in which the agent’s motion is observed. This affects both the heading of the swimmer and its linear velocity vector. The observation is updated accordingly by adjusting the front tip’s angle and rotating the $(x, y)$ velocity components.

\textbf{\textsc{RotatedHalfCheetah}.}
This task requires the agent to run forward in a planar environment. The original environment operates in a 2D plane. We perform a $45^\circ$ rotation on the $(x, z)$ components of the agent's torso position and velocity. The joint states and angular velocities remain unchanged.

\textbf{\textsc{RotatedWalker2d}.}
This task requires the agent to walk forward while maintaining balance. Similar to HalfCheetah, we rotate the observation of the agent’s torso height and linear velocity in the $(x, z)$ plane by $45^\circ$. This alters how progress and vertical motion are perceived, while the actuator and sensor states are kept intact.

\textbf{\textsc{RotatedAnt}.}
In this task, the agent is a quadruped and must navigate a 2D surface using four legs. We rotate the global $(x, y)$ velocity of the torso by $45^\circ$. When present, external contact forces are also rotated in the same plane. The result is a consistent shift in perceived motion direction across the ant’s high-dimensional observation space.

\textbf{Sensor Fusion.} 
In the fusion environments, instead of observing a $d_S$-dimensional observation vector, the agent receives a $10 \times d_S$-dimensional observation. More concretely, for each source observation element $i$, the target task provides $10$ measurements of the form: $x_i^{(j)} = w_j \cdot x_i + c_j$, for $j \in [10]$, where the weights $w_j$ sum to $1$ and the offsets $c_j$ sum to $0$. This simulates a sensor fusion scenario in which the undo map $U_\star$ sums the $10$ measurements corresponding to each original dimension to recover the source observation.

\textbf{\textsc{PolarHalfCheetah}.}
Here, we change the coordinate system of selected observation dimensions from Cartesian to polar coordinates. In particular, pairs of state variables such as the angular velocities of the back thigh and shin are transformed so that their joint $(x, y)$ representation becomes $(r, \theta)$, where $r = \sqrt{x^2 + y^2}$ and $\theta = \arctan2(y, x)$. This alters how the agent perceives limb velocities.

\textbf{\textsc{PolarAnt}.}
In this task, the angular velocity components of the torso along the $x$ and $y$ axes are represented in polar form. This affects how rotational motion is encoded.

\textbf{\textsc{PixelCartPole}.}
In this task, the agent receives pixel observations instead of low-dimensional sensor states. We extract a cropped $40 \times 60$ image centered around the cart, and the observation is the difference between two consecutive frames to capture dynamics. Since the cart’s absolute position and velocity cannot be inferred from pixels alone, we mask those variables from the source task to ensure comparability. This setting, inspired by \citet{lit_tl_1}, evaluates whether our method can transfer across a modality shift from sensor states to pixels.

\textbf{\textsc{ColorBreakout} and \textsc{ColorSpaceInvaders}.}
In these tasks, the source provides grayscale observations while the target provides RGB.

\textbf{\textsc{PixelBreakout}.}
In this task, the source provides semantic bitmaps corresponding to the ball, bricks, paddle, and trail, whereas the target provides grayscale pixel observations.
\section{Implementation Details}\label{sec:app.impl_details}

In our implementation, if an episode terminates early, we reset the environment but continue the action sequence indexing. This ensures that we can always collect $H^\dagger$ steps from an action sequence. The action sequences used to compute the state feature statistics are either randomly sampled or derived from the source policy $\pi_S$. Concretely, we perform a rollout of $\pi_S$ in the source MDP and record the action sequence taken by the policy. We then use this recorded sequence. One can potentially include both kinds of action sequences, which makes this procedure task-independent and effectively combines the benefits of both sampling strategies.

\textbf{Rotation and Fusion Tasks.} The undo map is a single linear layer, and we use {\alggen} with $M{=}1$ (learning rate $0.001$, weight decay $0.0001$, $10{,}000$ epochs). For \textsc{Swimmer} and \textsc{Walker2d}, we use randomly sampled action sequences of lengths 8 and 17, respectively, with $200$ and $100$ source trajectories and $3$ target trajectories per sequence. For \textsc{HalfCheetah} and \textsc{Ant}, the action sequences are obtained from the source policy (with lengths $17$ and $27$, respectively, and $100$ and $60$ source trajectories and $3$ target trajectories per sequence). We use up to $600$ action sequences.

\textbf{Polar Tasks.} The undo map is an MLP with two hidden layers ($64$ units per layer), and we use {\alggen} with $M{=}1$ (learning rate $0.001$, weight decay $0.0001$, $10{,}000$ epochs). We obtain $100$ action sequences of length $100$ from the source policy. Target statistics are computed from up to $10$ (\textsc{PolarHalfCheetah}) and $20$ (\textsc{PolarAnt}) trajectories per sequence.

\textbf{Sensor-to-Pixel Task.} The undo map is an MLP with one hidden layer ($16$ units), and we use {\alggen} with $M{=}2$ (learning rate $0.001$, weight decay $0.0001$, $10{,}000$ epochs). We use $50$ randomly sampled action sequences of length $50$. Target statistics are estimated with up to $20$ trajectories per sequence.

\textbf{Color Tasks.} The undo map is a single linear layer, and we use {\alggen} with $M{=}1$ (learning rate $0.01$, $\ell_1$ penalty of $0.001$, $3{,}000$ epochs). We use $10$ sequences of length $300$ (\textsc{ColorBreakout}) or $100$ (\textsc{ColorSpaceInvaders}), obtained from the source policy. Target statistics are estimated with up to $50$ (\textsc{ColorBreakout}) and $100$ (\textsc{ColorSpaceInvaders}) trajectories per sequence.

\textbf{Bitmaps-to-Pixel Task.} The undo map is an MLP with one hidden layer ($2048$ units), and we use {\alggen} with $M{=}2$ (learning rate $0.001$, weight decay $0.0001$, $2{,}000$ epochs). Here, we use $20$ source-policy action sequences of length $200$, with up to $50$ target trajectories per sequence.

More details are provided in the codebase. For the transfer baselines, we use the default hyperparameters provided in the original codebases.

\textbf{Compute resources.} All experiments were conducted on nodes of a shared cluster with Intel Xeon Gold 6426Y CPUs (up to $4.1$ GHz, $250$ GB RAM per node) and NVIDIA L40S GPUs ($46$ GB VRAM). Each experiment took approximately $5$ to $15$ minutes per seed.
\section{Robustness Analysis}\label{sec:app.robustness}

We conduct additional robustness experiments on \textsc{PixelCartPole} using five random seeds. \textsc{PixelCartPole} is a nontrivial transfer setting because the source policy operates on low-dimensional sensor observations, whereas the target provides only high-dimensional rendered pixel observations.

We consider two stochastic observation corruptions for which an exact deterministic (memoryless) undo map generally does not exist.
\begin{enumerate}[leftmargin=*]
    \item \textbf{Random Pixel Masking:} Each target pixel is independently set to zero with probability $p_m$.
    \item \textbf{Sticky Frames:} The previous observation is emitted instead of the current with probability $p_s$.
\end{enumerate}

Table~\ref{tab:robustness} reports the results at multiple target-sample budgets for the individual corruptions, as well as for a combined stochastic corruption consisting of pixel masking [$p_m = 0.05$] and sticky frames [$p_s = 0.01$]. All results are averaged over five random seeds.

\begin{table*}[h]
    \caption{Average return (standard deviation over five random seeds) on \textsc{PixelCartPole} under stochastic observation corruptions at different target-sample budgets.}
    \label{tab:robustness}
    \centering
    \scalebox{0.85}{
        \setlength\tabcolsep{5pt}
        \renewcommand{\arraystretch}{1.12}
        \begin{tabular}{lcccc}
            \toprule
            & \multicolumn{4}{c}{\textbf{Target Samples}} \\
            \cmidrule(lr){2-5}
            \textbf{Setting} & 12.5k & 25k & 37.5k & 50k \\
            \midrule
            Exact transformation
            & $86.12 \pm 18.20$ & $114.20 \pm 6.31$
            & $115.08 \pm 5.85$ & $118.96 \pm 3.82$ \\
            Pixel masking [$p_m = 0.05$]
            & $79.08 \pm 11.51$ & $99.96 \pm 13.08$
            & $99.52 \pm 7.00$ & $107.28 \pm 13.02$ \\
            Pixel masking [$p_m = 0.10$]
            & $70.00 \pm 20.55$ & $76.64 \pm 9.76$
            & $87.56 \pm 8.78$ & $91.36 \pm 13.63$ \\
            Sticky frames [$p_s = 0.01$]
            & $90.56 \pm 9.67$ & $95.76 \pm 16.62$
            & $104.12 \pm 9.05$ & $111.76 \pm 5.75$ \\
            Sticky frames [$p_s = 0.10$]
            & $48.04 \pm 13.28$ & $50.80 \pm 14.76$
            & $49.40 \pm 9.05$ & $57.68 \pm 10.52$ \\
            Combined [$p_m = 0.05$, $p_s = 0.01$]
            & $74.04 \pm 17.10$ & $91.92 \pm 27.11$
            & $98.88 \pm 13.22$ & $98.80 \pm 19.18$ \\
            \bottomrule
        \end{tabular}
    }
\end{table*}

Under $5\%$ pixel masking and $1\%$ sticky frames, {\alggen} remains close to its performance under the exact transformation. Performance decreases as the corruption becomes more severe with $10\%$ pixel masking or sticky frames. Sticky frames lead to a larger degradation because the entire current observation is occasionally replaced by stale information.

The results show that performance generally improves with additional target samples. Combining the two (mild) stochastic corruptions reduces performance relative to individual corruptions. However, it does not result in a drastic degradation of performance; {\alggen} reaches an average return of $98.80 \pm 19.18$ after 50k samples from the target.

Overall, these results show that {\alggen} remains effective under moderate stochastic observation corruption and degrades as the exact undo-map assumption is increasingly violated.
}

\end{document}